\documentclass[letterpaper, 10 pt, conference]{ieeeconf}  

\IEEEoverridecommandlockouts                              

\usepackage{graphics} 
\graphicspath{ {figures/} }
\usepackage{caption}
\usepackage{subcaption}
\usepackage{epsfig} 
\usepackage{amsmath} 
\usepackage[ruled]{algorithm2e} 
\usepackage{url}
\usepackage{bm} 
\usepackage{comment}
\usepackage{amssymb}
\usepackage{multirow}
\usepackage{tabularx}
\usepackage{cite}
\newcolumntype{C}[1]{>{\centering\arraybackslash}p{#1}}

\title{\LARGE \bf
Model-Based Manipulation of Linear Flexible Objects with Visual Curvature Feedback
}

\author{Peng Chang$^{1}$ and Ta\c{s}k{\i}n~Pad{\i}r$^{2}$
\thanks{$^{1}$Department of Electrical and Computer Engineering, Northeastern University, Boston, Massachusetts, USA. {\tt\small chang.pe@husky.neu.edu}
        $^{2}$Institute for Experiential Robotics, Northeastern University, Boston, Massachusetts, USA. {\tt\small t.padir@northeastern.edu}
        }%
}

\begin{document}

\maketitle
\thispagestyle{empty}
\pagestyle{empty}

\begin{abstract}

Manipulation of deformable objects is a desired skill in making robots ubiquitous in manufacturing, service, healthcare, and security. Deformable objects are common in our daily lives, e.g., wires, clothes, bed sheets, etc., and are significantly more difficult to model than rigid objects. In this study, we investigate vision-based manipulation of linear flexible objects such as cables. We propose a geometric modeling method that is based on visual feedback to develop a general representation of the linear flexible object that is subject to gravity. The model characterizes the shape of the object by combining the curvatures on two projection planes. In this approach, we achieve tracking of the position and orientation (pose) of a cable-like object, the pose of its tip, and the pose of the selected grasp point on the object, which enables closed-loop manipulation of the object. We demonstrate the feasibility of our approach by  completing the Plug Task used in the 2015 DARPA Robotics Challenge Finals, which involves unplugging a power cable from one socket and plugging it into another. Experiments show that we can successfully complete the task autonomously within 30 seconds. 

\end{abstract}

\section{INTRODUCTION}

Manipulation of linear flexible objects is of great interest in many applications including manufacturing, service, health and disaster response. Indeed, the topic has been the focus of many research studies in recent years. Ropes, clothes, organs and cables are common flexible objects used in deformable object manipulation \cite{sanchez2018robotic, lee2015learning, nair2017combining, miller2012geometric, mallapragada2011toward, white1988remotely}.

Cables are linear flexible objects that are common in industrial 
and domestic environments. The 2015 DARPA Robotics Challenge (DRC) Finals was aimed at advancing the capabilities of human-robot teams in responding to natural and man-made disasters. The Plug Task in this challenge required the robot pulling a power cable out of a socket and plug it into another \cite{spenko2018darpa}. Among six teams that completed the task, Team WPI-CMU had the fastest completion time with 5 minutes and 7 seconds \cite{cisneros2016effective}. Their approach used teleoperation by the human operator to grasp the plug and to complete the task \cite{dedonato2017team}. The other five teams also did not automate this task, and they took longer times to perform the task. 


In most of the related literature of manipulating cable-like objects, the physical model of the deformed object needs to be known before the manipulation such that the deformation of the object can be predicted. Chen and Zheng \cite{chen1992deformation} used a cubic spline function to estimate the contour of a flexible aluminum beam in the 2-D plane, and a non-linear model was applied to describe the deformation based on the material characteristics identified by vision sensors. A systematic method to model the bend, twist, and extensional deformations of flexible linear objects was presented in \cite{wakamatsu1995modeling}, and the stable deformed shape of the flexible object was characterized by minimizing the potential energy under the geometric constraints. Nakagaki et al. \cite{nakagaki1996study} extended this work and proposed a method to estimate the force on a wire based on its shape observed by stereo vision. This method was employed for an insertion task of a flexible wire into a hole. Caldwell et al. \cite{caldwell2014optimal} presented a technique to model a flexible loop by a chain of rigid bodies connected by torsional springs. Yoshida et al. \cite{yoshida2015simulation} proposed a method for planning motions of a ring-shaped object based on precise simulation using Finite Element Method (FEM). These work are based on a priori knowledge of the deformation model of flexible objects. However, for different cables with different thickness, material, or function, the physical and deformation models vary substantially. So, these methods are difficult to be generalized to different cable-like objects.

\begin{figure}%
\vspace{2mm}
\centering
\begin{subfigure}[b]{0.475\linewidth}
\label{fig:intro-a}%
\includegraphics[width=\textwidth, height=.7\textwidth]{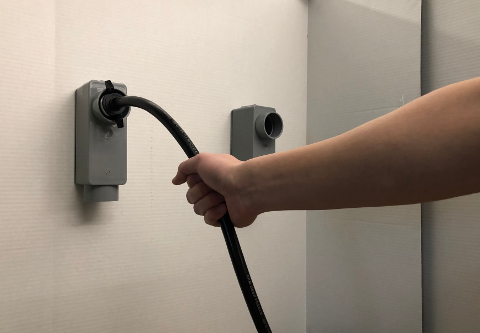}%
\caption[a]{}
\end{subfigure}
\begin{subfigure}[b]{0.475\linewidth}
\label{fig:intro-b}%
\includegraphics[width=\textwidth, height=.7\textwidth]{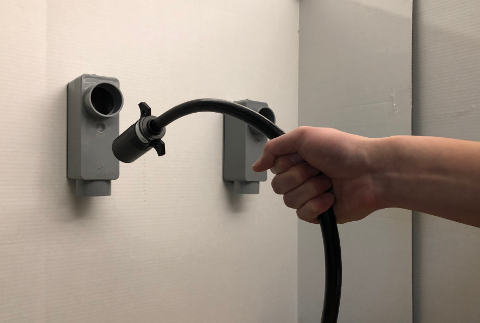} 
\caption[b]{}
\end{subfigure} \\
\begin{subfigure}[b]{0.475\linewidth}
\label{fig:intro-c}%
\includegraphics[width=\textwidth, height=.7\textwidth]{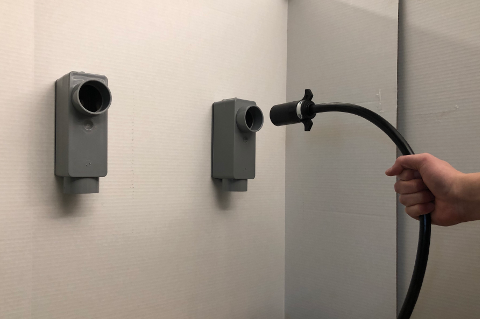}%
\caption[c]{}
\end{subfigure}
\begin{subfigure}[b]{0.475\linewidth}
\label{fig:intro-d}%
\includegraphics[width=\textwidth, height=.7\textwidth]{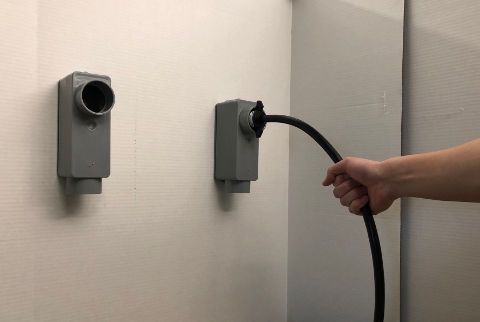}%
\caption[d]{}
\end{subfigure}
\caption{Manipulation of a power cable by humans: the goal of the task is to switch the plug from one socket to another.}%
\label{fig:intro}%
\vspace{-2mm}
\end{figure}

There are also approaches which are not based on the physical model of the flexible object, and our method falls into this category. Navarro-Alarcon et al. \cite{navarro2013model, navarro2014visual, navarro2016automatic} proposed a framework for automatic manipulation of deformable objects with an adaptive defomation model in 3-D space. Cartesian features composed of points, lines, and angles have been used to represent the deformation. Recently, Navarro-Alarcon and Liu \cite{navarro2018fourier} proposed a representation of the object's shape based on a truncated Fourier series, and this model allows the robotic arm to deform the soft objects to desired contours in 2-D plane. Recently, Zhu et al. \cite{zhu2018dual} extended the work in \cite{navarro2018fourier} and used the Fourier-based method to control the shape of a flexible cable in 2-D space. Compared with the Fourier-based visual servoing method, the SPR-RWLS method proposed by Jin et al. \cite{jin2019robust} took visual tracking uncertainties into consideration and showed robustness in the presence of outliers and occlusions for cable manipulation.

Inspired by the existing work, we propose a method for geometrical modeling of general linear flexible objects. Moreover, we use this method to systematically complete the DRC Plug Task. In order to detect and track customized small objects, such as cables, in 3-D space, \cite{de2018integration} used a neural network-based detection method, and we extend their detection and tracking methods by introducing color detection into the loop. This paper has the following contributions: (1) it introduces a novel feedback characterization of linear flexible objects subject to gravity by combining curvatures on two projection planes which enables tracking of the position and orientation (pose) of a cable-like object, the pose of its tip, and the pose of the selected grasp point on the object, and a robust pose alignment controller which has the ability to control the shape of the object; and (2) it presents an autonomous system framework for accomplishing the DRC Plug Task. It should be noted that DRC Plug Task has been selected as the validation study in this research for two reasons: (i) Our team participated in the DRC Finals \cite{dedonato2017team}, and (ii) The DRC Plug Task provides a well-defined set of requirements to generalize the approach presented here to other applications.

This paper is organized as follows. Section II presents the problem statement and system framework. Section III describes the methodology of obtaining a reliable geometrical model of the cables on-the-fly and a robust pose alignment controller. In Section IV, we show the feasibility of our method experimentally and discuss the experimental results. Section V includes the conclusion and directions for future work.

\section{PROBLEM STATEMENT AND TASK DESCRIPTION} \label{section:system_pipeline}
\begin{figure}[ht]
    \centering
    \includegraphics[width=8cm]{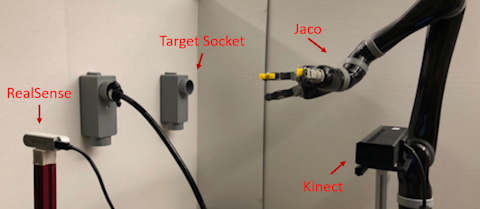}
    \caption{DRC Plug Task setup.}
    \label{fig:workspace}
    \vspace{-4mm}
\end{figure}

\begin{figure}[ht]%
\centering
\begin{subfigure}[b]{0.45\linewidth}
\label{fig:finger-a}%
\includegraphics[width=\textwidth, height=.8\textwidth]{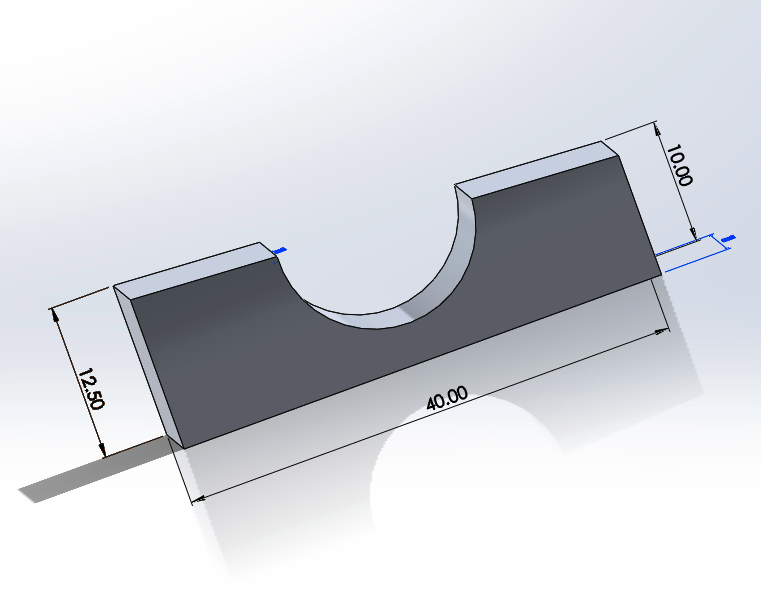}%
\caption[a]{Gripper attachment design in SolidWorks.}
\end{subfigure}
\begin{subfigure}[b]{0.45\linewidth}
\label{fig:finger-b}%
\includegraphics[width=\textwidth, height=.8\textwidth]{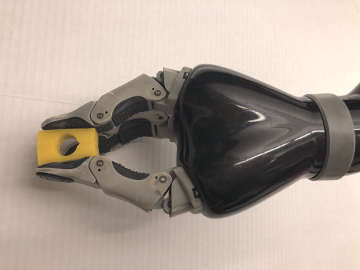} 
\caption[b]{Attachments mounted on the robot fingers.}
\end{subfigure}
\caption{Finger attachment design.}%
\label{fig:finger}%
\vspace{-2mm}
\end{figure}

The DRC Plug Task involves unplugging a power cable from one socket and plugging it into another. The problem involves detecting a flexible cable in a noisy environment, grasping the cable with a feasible grasp pose, controlling the shape of the cable to match with the target pose, and inserting the cable-tip to the target socket. A benchmarking setup is used in this study shown in Fig.~\ref{fig:workspace}. The setup consists of a power cable (Deka Wire DW04914-1) with a plug (Optronics A7WCB) at the tip, two power sockets (McMaster-Carr 7905k35) attached on the wall, two neodymium disk magnets (DIYMAG HLMAG03) glued to the socket and the plug to provide a suction force \cite{spenko2018darpa}. Our system framework consists of a 6-DOF JACO v2 arm with three fingers, and two RGB-D cameras, i.e., Realsense D435 and Microsoft Kinect. The Realsense camera is used to model and track the shape of the cable, while the Kinect camera is used to estimate the socket pose and filter the point cloud. This scenario corresponds to a humanoid robot equipped with a depth camera at its left arm wrist and another depth camera at its head. One arm performing the task while the other is providing the perception feedback. In addition, a custom attachment is mounted on the fingers of the gripper to enable a form-closure grasp around the cable cross section. Figure~\ref{fig:finger} shows this attachment in SolidWorks and mounted on the fingers.

\begin{figure*}[ht]
    \vspace{2mm}
    \centering
    \includegraphics[width=.7\linewidth]{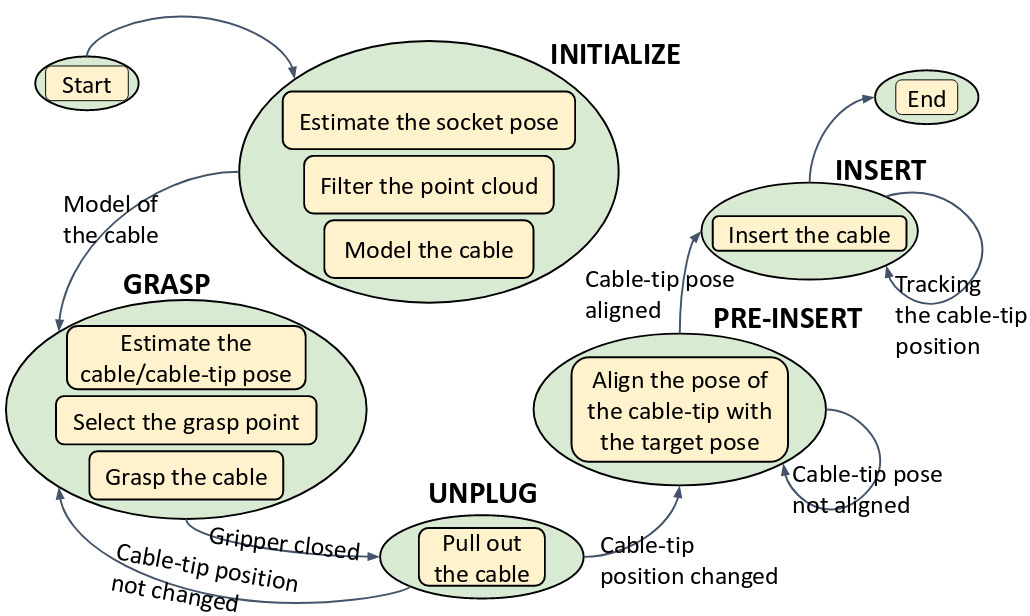}
    \caption{Overall computational model of our system can be represented in a event-driven state machine.}
    \label{fig:flowchart}
    \vspace{-2mm}
\end{figure*}

From a system design perspective, in order to complete the task autonomously, the system needs to be able to model a linear flexible object, to keep track of the cable configuration and to detect the pose of the target socket through pose estimation, to plan motions to move the robot to desired configurations, and to align the cable-tip with the target socket by a controller. A flow of the proposed system architecture is presented in Fig.~\ref{fig:flowchart}.

Similar to the one-handed cable manipulation by a human (Fig.~\ref{fig:intro}), we can divide the entire process into five phases: \textsc{Initialize}, \textsc{Grasp}, \textsc{Unplug}, \textsc{Pre-insert}, and \textsc{Insert}. Our goal is to give the robot the ability to automatically switch the phases. From the \textsc{Initialize} phase to the \textsc{Grasp} phase, the system needs to sense the environment, including estimating the socket pose, filtering the point cloud, and modeling the cable. After getting the model of the cable, the system will estimate the poses of the cable and cable-tip, and select the grasp point. Then the robot will go and \textsc{grasp} the cable. From the \textsc{Unplug} phase to the \textsc{Pre-insert} phase, the robot needs to align the pose of the cable-tip with the target pose. From the \textsc{Grasp} phase to the \textsc{Unplug} phase, a planned motion perpendicular to the target socket hole plane and away from the target socket is executed by the robot, and another planned motion perpendicular to the socket hole plane but close to the socket is executed by the robot in order to transfer from the \textsc{Pre-insert} phase to the \textsc{Insert} phase.

\section{METHODOLOGY} \label{section:methodology}
Our approach relies on five main robot-centric capabilities: (1) Target pose estimation, (2) Real-time point cloud filtering, (3) Modeling of a linear flexible object, (4) Cable/cable-tip pose estimation, (5) Pose alignment controller.
\subsection{Target pose estimation} \label{s1.1}
\begin{figure}[htp]
    \vspace{2mm}
    \centering
    \includegraphics[height=.3\linewidth]{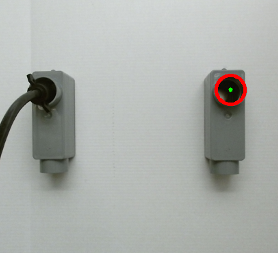}\hfill
    \includegraphics[height=.3\linewidth]{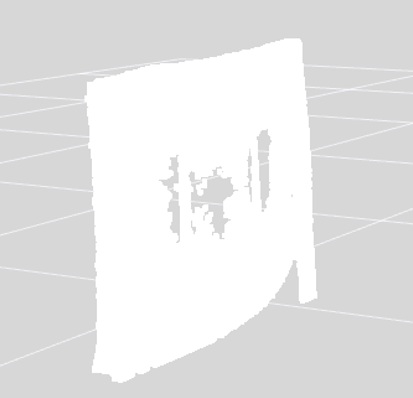}\hfill
    \includegraphics[height=.3\linewidth]{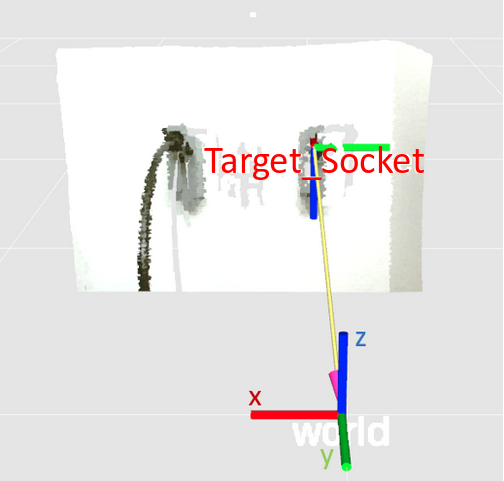}
    \caption{Steps of socket pose estimation: (left) hough circle detection, (middle) plane detection, (right) resulting socket hole frame.}
    \label{fig:Socket_Pose_Estimation}
    \vspace{-2mm}
\end{figure}

The target socket in Fig.~\ref{fig:workspace} is the final goal for the cable-tip. We used the Hough Circle Transform \cite{yuen1990comparative} to detect the hole of the target socket, as shown in Fig.~\ref{fig:Socket_Pose_Estimation} (left). The center of the detected red circle is marked by a green dot and we denote the coordinate of the center as ($x_{target},y_{target}$). If we denote the 3-D positions of the target socket center as ($X_{target},Y_{target},Z_{target}$), we can get $Z_{target}$ directly by reading the depth information at the coordinates ($x_{target},y_{target}$) from the Kinect camera, and ($X_{target},Y_{target}$) can be calculated by using the perspective projection: 
\begin{equation} \label{eq:perspective_projection}
x_{target}=f*\frac{X_{target}}{Z_{target}},\quad y_{target}=f*\frac{Y_{target}}{Z_{target}},
\end{equation}
where \(f\) is the focal length of the camera. 
In order to get the orientation of the target socket hole, the orientation of vector that is normal to the wall surface is used since the target socket hole plane is parallel to the wall. Random sample consensus (RANSAC) method \cite{fischler1981random} is applied to detect the wall plane from the point cloud that is obtained from Kinect camera, as seen in Fig.~\ref{fig:Socket_Pose_Estimation} (middle). We can then find the normal vector to the wall plane. And the $x$ axis of the target socket frame is set along this normal vector. Because the target socket hole has the round shape, we can pick any rotation angle about the $x$ axis (we set it to 0) to define the target socket frame. The corresponding frame (named as ``Target\_Socket'') is published through ROS, as shown in Fig.~\ref{fig:Socket_Pose_Estimation} (right).
\subsection{Real-time point cloud filtering}\label{s1.2}

\begin{figure}[ht]
    \vspace{2mm}
    \centering
    \includegraphics[height=.25\linewidth, width=.3\linewidth]{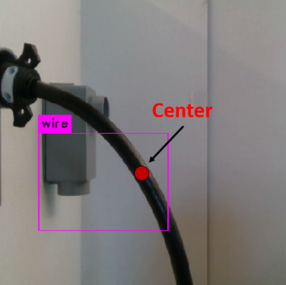}\hfill
    \includegraphics[height=.25\linewidth, width=.3\linewidth]{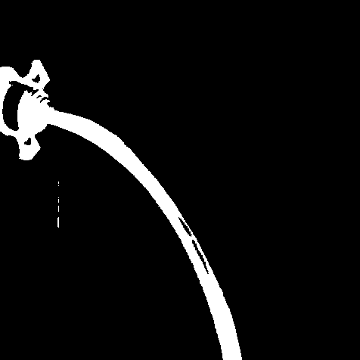}\hfill
    \includegraphics[height=.25\linewidth, width=.35\linewidth]{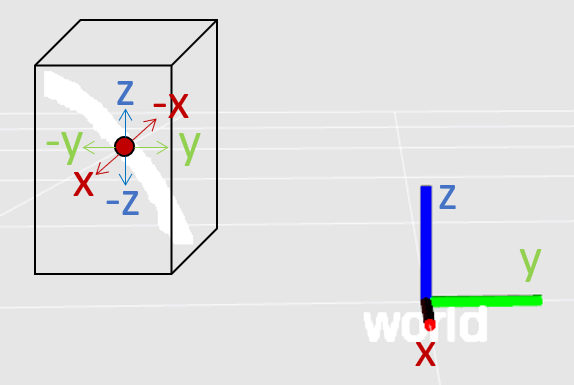}
    \caption{Steps for obtaining the filtered point cloud: (left) YOLO object detection, (middle) color detection, (right) point cloud filter.}
    \label{fig:Filtered_Point_Cloud}
    \vspace{-2mm}
\end{figure}

It is challenging to find specific irregular objects in a noisy environment. In this study, a real-time object detection method called YOLO \cite{redmon2016you} is used for detecting the flexible power cable in the 2-D images (Fig.~\ref{fig:Filtered_Point_Cloud} (left)). The process requires training a convolutional neural network (CNN) with self-labeled images. We trained the network for 40,000 steps with 100 labeled images. The output of YOLO is a bounding box that narrows the region down to a search for the cable. To get the pixels corresponding to the object, we then use the color information and detect the black pixels in this region (Fig.~\ref{fig:Filtered_Point_Cloud} (middle)). These pixels are stored in a set in the order from top to bottom and left to right, and we define the middle pixel as the center pixel of the object which is marked as ``Center'' in Fig.~\ref{fig:Filtered_Point_Cloud} (left). By using (\ref{eq:perspective_projection}) we mentioned above, we can get the 3-D positions of the object center. A \textit{PassThrough} filter from the point cloud library (pcl) \cite{rusu2011point} was used to filter the point cloud from three dimensions as shown in Fig.~\ref{fig:Filtered_Point_Cloud} (right). The point cloud in the narrowed bounding box served to model the cable. It is published through ROS in a frequency of 30 $Hz$.   

\subsection{Modeling of a linear flexible object} \label{s1.3}
In Fig.\ref{fig:Filtered_Point_Cloud} (right), the remaining points in the filtered point cloud are shown in white with respect to the world frame. The $x$, $y$, and $z$ axes of the world frame are represented by the red, green, and blue bars, respectively. Modeling the cable in 3-D space is still a research problem in linear flexible object manipulation. On the other hand, modeling a curved line in 2-D space is a solved problem. We can exploit this by projecting the 3-D curve onto 2-D spaces. In this study, we project the 3-D point cloud onto the $y$-$x$ and $y$-$z$ planes. But the projection in 2D may have multiple corresponding $x$ or $z$ values to the $y$ coordinate. For example, if the cable has a ``$\supset$'' shape, in order to get an unique projection model, we filtered out the point cloud below the rightmost point. And we assume that the cable is not in more complex shapes, e.g., ``S'' or ``$\alpha$'' shape. Due to the relatively high stiffness and large bending radius of the power cables, we found that a quadratic polynomial equation is sufficient to represent their shapes:
\begin{equation} \label{eq:x_y}
x=a_0+a_1y+a_2y^2,
\end{equation}
\begin{equation} \label{eq:z_y}
z=b_0+b_1y+b_2y^2.
\end{equation}
The polynomial coefficients $a_0$, $a_1$, $a_2$, $b_0$, $b_1$, and $b_2$ are estimated by using the least squares method, and they continuously change based on the shape of the power cable. In order to model the cable in 3-D, we uniformly sample the points in Fig.~\ref{fig:Filtered_Point_Cloud} (right) along the $y$-axis. The corresponding $x$ and $z$ values can be calculated by using (\ref{eq:x_y}) and (\ref{eq:z_y}). This projection method is efficient, and the model can be published through ROS in a frequency of 29.997 $Hz$.

\begin{figure}[ht]
    \centering
    \includegraphics[width=.5\linewidth]{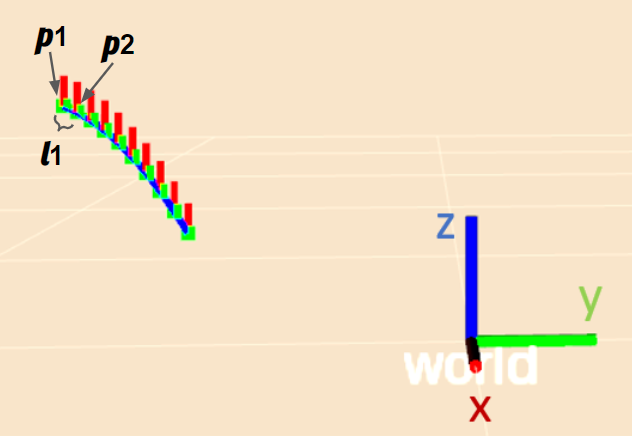}
    \caption{A modeling example with 10 sample points.}
    \label{fig:modeling_example}
    \vspace{-2mm}
\end{figure}

Figure~\ref{fig:modeling_example} shows an example modeling result with 10 sample points. The power cable is visualized in Rviz by markers. Green markers represent the sampling points, consecutive points are connected by blue lines, and vertical upward red lines are used to illustrate the displacements between the points. 

\subsection{Cable/cable-tip pose estimation} \label{s1.4}

\begin{figure}[ht]%
\vspace{2mm}
\centering
\begin{subfigure}[b]{0.4\linewidth}
\label{fig:wire_tip-a}%
\includegraphics[width=\textwidth, height=.6\textwidth]{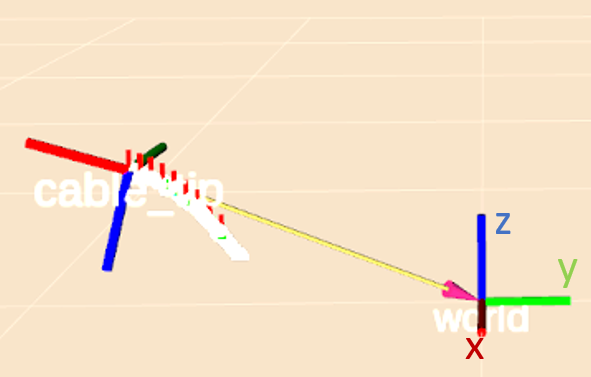}%
\end{subfigure}
\begin{subfigure}[b]{0.4\linewidth}
\label{fig:wire_tip-b}%
\includegraphics[width=\textwidth, height=.6\textwidth]{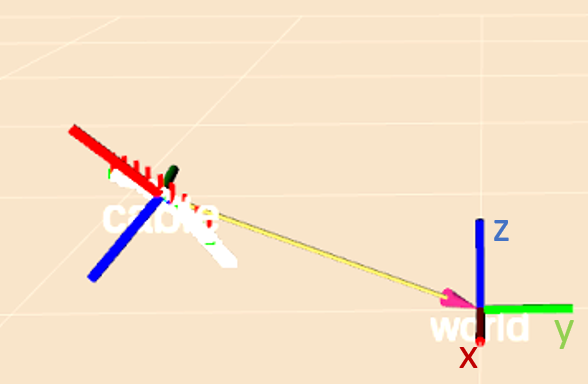} 
\end{subfigure} \\
\caption{(left) cable-tip frame, (right) cable frame.}%
\label{fig:wire/wire_tip_pose_estimation}%
\end{figure}

In order for the robot to detect and track the location of the cable and cable-tip, we propose a piece-wise linear model for the object. In Fig.~\ref{fig:modeling_example}, we denote the leftmost point as $p_1$ and the remaining points as $p_2,p_3,...,p_N$ from left to right, where $N$ is the number of sample points. We denote the leftmost blue line that connects $p_1$ and $p_2$ as $l_1$ and rest of the blue lines as $l_2,l_3,...,l_{N-1}$. We assume that the line $l_i$ is the tangent at the point $p_i$. Because the cable-tip has the round shape, its roll angle can be neglected (we set it to 0). After setting the point $p_1$ as the origin and defining the $x$ axis to the tangent of the point $p_1$, the ``cable\_tip'' frame is defined in Fig.~\ref{fig:wire/wire_tip_pose_estimation} (left).

Now that the positions of all the sample points and the pose of the cable-tip are known with respect to the robot frame, a feasible grasp pose for the robot can be calculated to \textsc{grasp} the cable. Because the plug is inside the socket, we consider only the sampling points on the cable except the cable-tip as the grasp point candidates. To select the grasp point on the cable, the following trade off needs to be taken into account. As the grasp point gets closer to the cable-tip, the risk of the robot colliding with the socket or the wall increases and the number of points in the filtered point cloud reduces. On the other hand, when the grasp point is too far away from the cable-tip, the robot can not align the cable-tip pose with the target socket pose because the cable will dangle substantially. Thus, the grasp point needs to be selected empirically to meet these two constraints. For this purpose, we pre-define a minimum distance from the grasp point to the cable-tip ($p_1$) along the cable as $d_{min}$, and a maximum distance as $d_{max}$. $d_{min}$ and $d_{max}$ are pre-defined with the robot holding the cable before the experiment. With this step, the cable deformation characteristics can be ignored for selecting the grasp point, and this vision-only method is satisfied for selecting a feasible grasp point. Since we can calculate the length of the lines $l_1,...,l_{N-1}$, we can also calculate the distance $d_{s}$ from each sampling point to the cable-tip by using: 
 \begin{equation} \label{eq:di}
d_{s} \triangleq \sum_{i=1}^{s} l_i, \  s=1,...,N-1.
\end{equation}
 The grasp point needs to be selected such that $d_{s}\in[d_{min},d_{max}]$ in order for the robot to be able to align the cable-tip pose with the target pose. By using the same method of estimating the ``cable\_tip'' frame, we can get the ``cable'' frame (Fig.~\ref{fig:wire/wire_tip_pose_estimation} (right)), which is used to get the grasp pose of the robot's end-effector. The grasp pose is then used as the target of the motion planner, which is MoveIt! \cite{chitta2012moveit} in our case.
 
\begin{figure}[ht]%
\centering
\begin{subfigure}[b]{0.4\linewidth}
\label{fig:wire_tip_pose_alignment-a}%
\includegraphics[width=\textwidth, height=.6\textwidth]{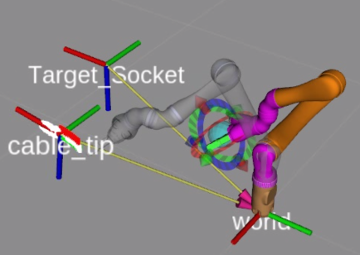}%
\caption[a]{Cable-tip pose after\\ unplugging.}
\end{subfigure}
\begin{subfigure}[b]{0.4\linewidth}
\label{fig:wire_tip_pose_alignment-b}%
\includegraphics[width=\textwidth, height=.6\textwidth]{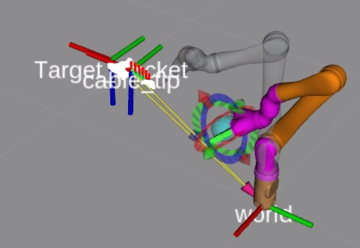} 
\caption[b]{Cable-tip pose after\\ pose alignment.}
\end{subfigure}
\caption{Cable-tip pose alignment.}%
\label{fig:wire_tip_pose_alignment}%
\vspace{-4mm}
\end{figure}

 Once grasping of the cable is completed, a pulling motion along the $y$ axis of the ``world'' frame is generated to \textsc{unplug} the cable. Figure~\ref{fig:wire_tip_pose_alignment} (a) shows the ``cable\_tip'' frame after the robot unplugs the power cable. We filter out the point cloud of the plug during modeling which improves the accuracy of the model because the plug is a straight, rigid object.

\subsection{Pose alignment controller} \label{s1.5}

In order to finish the \textsc{insertion} task, the robot needs to adjust the pose of the cable-tip to match with the pose of the target socket. A \textsc{pre-insert} frame right in front of the target socket is defined as our target pose for the pose alignment controller. 

To explain this algorithm in a succinct way, we denote the transformation matrix from the ``cable\_tip'' frame to the \textsc{pre-insert} frame as $\mathbf{T}^{pre}_{ct}$, the transformation matrix from the \textsc{pre-insert} frame to the ``end-effector'' frame as $\mathbf{T}^{ee}_{pre}$, the transformation matrix from the ``cable\_tip'' frame to the ``end-effector'' frame as $\mathbf{T}^{ee}_{ct}$.

$[\Delta x~\Delta y~\Delta z~\Delta \alpha~\Delta \beta~\Delta \gamma]^\top$ are the translational ($x$,~$y$,~$z$) and rotational ($roll$,~$pitch$,~$yaw$) deviations in the ``end-effector'' frame which can be calculated as

\begin{equation}
    \begin{bmatrix}
     \Delta x \\
     \Delta y \\ 
     \Delta z
    \end{bmatrix}
    =
    \begin{bmatrix}
     x_1 \\
     y_1 \\ 
     z_1
    \end{bmatrix}
    -
    \begin{bmatrix}
     x_2 \\
     y_2 \\ 
     z_2
    \end{bmatrix},
    \begin{bmatrix}
     \Delta \alpha \\
     \Delta \beta \\ 
     \Delta \gamma
    \end{bmatrix}
    =
    \begin{bmatrix}
     \alpha_1 \\
     \beta_1\\ 
     \gamma_1
    \end{bmatrix}
    -
    \begin{bmatrix}
     \alpha_2 \\
     \beta_2\\ 
     \gamma_2
    \end{bmatrix},
    \label{eq:translation_rotation}
\end{equation}

where $x_1$, $y_1$, $z_1$, $\alpha_1$, $\beta_1$, and $\gamma_1$ are translational and rotational terms from the transformation $\mathbf{T}^{ee}_{pre}$ in the ``end-effector'' frame, $x_2$, $y_2$, $z_2$, $\alpha_2$, $\beta_2$, and $\gamma_2$ are translational and rotational terms from the transformation $\mathbf{T}^{ee}_{ct}$ in the ``end-effector'' frame.

Then the linear and angular velocities in the ``end-effector'' frame can be calculated \cite{khalil2004modeling}:
\begin{equation}
    \begin{bmatrix}
     \dot{x} \\
     \dot{y} \\ 
     \dot{z}
    \end{bmatrix}
    =
    \begin{bmatrix}
     \Delta x \\
     \Delta y \\ 
     \Delta z
    \end{bmatrix}
    \frac{1}{\Delta t},
\end{equation}
\begin{equation}
    \begin{bmatrix}
     w_x \\
     w_y \\ 
     w_z
    \end{bmatrix}
    =
     \begin{bmatrix}
    1 & 0 &-sin(\Delta \beta)\\
    0 &cos(\Delta \alpha)  &cos(\Delta \beta)sin(\Delta \alpha)\\
    0 & -sin(\Delta \alpha)             &cos(\Delta \beta)cos(\Delta \alpha)
    \end{bmatrix}
    \begin{bmatrix}
    \Delta \alpha \\
    \Delta \beta \\
    \Delta \gamma 
    \end{bmatrix}
    \frac{1}{\Delta t},
\end{equation}
where $\Delta t$ is the execution time, $\dot{x}, \dot{y}, \dot{z}, w_x, w_y, and$ $w_z$ are linear and angular velocities in the ``end-effector'' frame, which will be used as inputs for our PD controller.

A PD controller denoted as pose alignment controller was designed to calculate the Cartesian velocity ($\dot{\mathbf{x}}$) for the robot's end-effector:
\begin{equation}
    \dot{\mathbf{x}} = K_p\mathbf{e} + K_d\dot{\mathbf{e}},
\end{equation}
where $\mathbf{e}$ is the error, which is $[\dot{x}~\dot{y}~\dot{z}~w_x~w_y~w_z]^\top$, $K_p$ and $K_d$ are proportional and derivative terms. Our designed values for $K_p$ and $K_d$ are 2.0 and 0.2.

After getting the Cartesian velocity from our PD controller, a velocity controller built in the Jaco robotic arm generates the controlled joint velocity ($\dot{\mathbf{q}}$) by using robot Jacobian matrix ($\mathbf{J}$):
\begin{equation}
    \dot{\mathbf{q}}=\mathbf{J}^{-1}\dot{\mathbf{x}}.
\end{equation}

Algorithm \ref{algorithm:pose_alignment} shows the control loop. The control loop keeps running until the translational deviations $(\Delta x_e$, $\Delta y_e$, $\Delta z_e)$ and the rotational deviations $(\Delta \alpha_e$, $\Delta \beta_e$, $\Delta \gamma_e)$ from the ``cable\_tip'' frame to the \textsc{pre-insert} frame ($\mathbf{T}^{pre}_{ct}$) satisfy the thresholds which are $\epsilon_1$ ($m$) and $\epsilon_2$ ($rad$). A successful example of the robot aligning the cable to \textsc{pre-insert} frame is shown in Fig.~\ref{fig:wire_tip_pose_alignment} (b). For the final \textsc{insertion} step, a translation along the $x$ axis of the ``cable\_tip'' frame is applied. The insertion is facilitated by the magnets on the plug and in the socket. 
\begin{algorithm}[]
 \KwResult{``cable\_tip'' frame = \textsc{pre-insert} frame}
 \textbf{Calculation:\{}\\
 Calculate $\mathbf{T}^{pre}_{ct}$, $\mathbf{T}^{ee}_{pre}$, $\mathbf{T}^{ee}_{ct}$;\\
 Get the translational ($\Delta x_e,\Delta y_e,\Delta z_e$) and rotational ($\Delta \alpha_e,\Delta \beta_e,\Delta \gamma_e$) differences of the transformation; \\
 Get the translational and rotational differences ($\Delta x,\Delta y,\Delta z,\Delta \alpha,\Delta \beta,\Delta \gamma$) in the ``end-effector'' frame ;\\
 Calculate the linear and angular velocities ($\dot{x}, \dot{y}, \dot{z}, w_x, w_y, w_z$) for the pose alignment controller; \}\\
 \While{PD controller: $max(\Delta x_e,\Delta y_e,\Delta z_e)\geq \epsilon_1$ $\lor$ \\ $max(\Delta \alpha_e,\Delta \beta_e,\Delta \gamma_e) \geq \epsilon_2$}{
 Pose alignment controller generates Cartesian velocity ($\dot{\mathbf{x}}$);\\
 Velocity controller takes the Cartesian velocity and generate velocities for robot joints ($\dot{\mathbf{q}}$);\\
 The cable-tip moves towards the \textsc{pre-insert} pose by the robot's end-effector;\\
 \textbf{Calculation};\\
}
 \caption{Pose alignment algorithm}
 \label{algorithm:pose_alignment}
\end{algorithm}
\section{EXPERIMENTAL VALIDATION} \label{section:experiment}

Our system framework is evaluated by running the DRC Plug Task for 20 trials. The minimal, maximum, and average completion time are 29.3823~$s$, 33.9710~$s$, and 31.5288~$s$.

In order to validate the performance of our modeling method and pose alignment controller, we implement three experiments. The first experiment is for testing the grasp functionality with different initial poses of the cable. The second experiment shows the performance of the pose alignment controller with two different cables. The last experiment shows the robustness of our pose alignment controller under external disturbances.

\begin{figure}[ht]
    \centering
    \includegraphics[height=.2\linewidth, width=.3\linewidth]{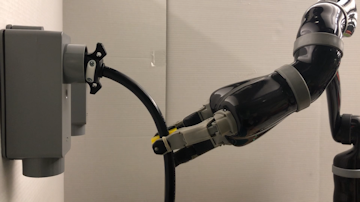}\hfill
    \includegraphics[height=.2\linewidth, width=.3\linewidth]{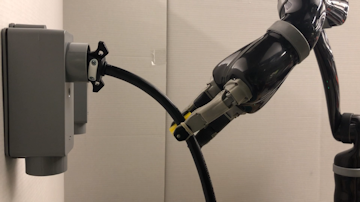}\hfill
    \includegraphics[height=.2\linewidth, width=.3\linewidth]{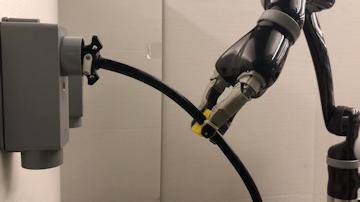}
    \caption{Experiment 1: The robot successfully grasps the cable with three different initial states.}
    \label{fig:different_initial_cable}
    \vspace{-4mm}
\end{figure}

Recalling the trade-off mentioned in Section \ref{s1.4}, we need to pick values of $d_{min}$ and $d_{max}$ for different cables. For the power cable (with plug), we define $d_{min}$ = 18~$cm$ and $d_{max}$ = 30~$cm$. For the HDMI cable, these parameters are selected as $d_{min}$ = 12~$cm$ and $d_{max}$ = 24~$cm$. Our method will prioritize the sample point closest to the midpoint in this range as grasp point when performing tasks. Figure~\ref{fig:different_initial_cable} shows the robot is able to grasp the cable with three different initial states. Table~\ref{table:model_parameters} shows the model parameters for these states.
\begin{table}[ht]
\centering
\caption{Estimated model parameters of the power cable for three different initial states.}
\label{table:model_parameters}
\begin{tabular}{|c|c|c|c|c|c|c|}
\hline
Poses    & $a_0$     & $a_1$     & $a_2$     & $b_0$     & $b_1$      & $b_2$      \\ \hline
left          & -1.882 & -7.295 & -5.773 & -6.041 & -20.004 & -15.528  \\ \hline
middle        & -0.385 & -2.670 & -2.191 & -1.917 & -6.981  & -5.243  \\ \hline
right         & 0.006  & -1.377 & -1.125 & -0.979 & -3.955  & -2.806 \\ \hline
\end{tabular}
\end{table}

\begin{figure}[ht]%
\vspace{2mm}
\centering
\begin{subfigure}[b]{0.33\linewidth}
\label{fig:power_a}%
\includegraphics[width=.8\textwidth, height=.5\textwidth]{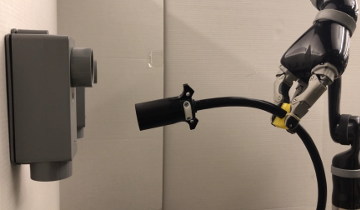}%
\caption[a]{Power cable initial configuration.}
\end{subfigure}
\begin{subfigure}[b]{0.33\linewidth}
\label{fig:power_b}%
\includegraphics[width=.8\textwidth, height=.5\textwidth]{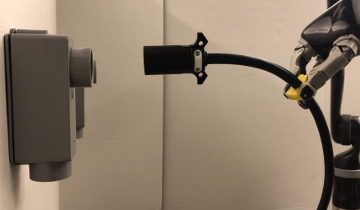}
\caption[b]{Power cable final configuration.}
\end{subfigure}
\begin{subfigure}[b]{0.33\linewidth}
\label{fig:hdmi_a}%
\includegraphics[width=.8\textwidth, height=.5\textwidth]{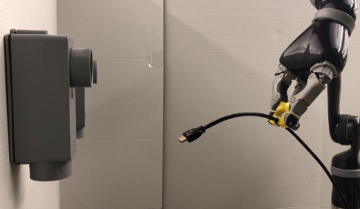}%
\caption[a]{HDMI cable initial configuration.}
\end{subfigure}
\begin{subfigure}[b]{0.33\linewidth}
\label{fig:hdmi_b}%
\includegraphics[width=.8\textwidth, height=.5\textwidth]{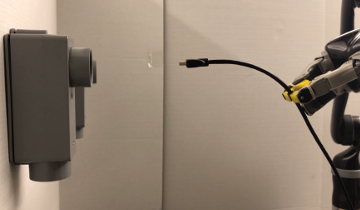}
\caption[b]{HDMI cable final configuration.}
\end{subfigure}
\caption{Experiment 2: Power/HDMI cable pose alignment.}
\label{fig:pose_alignment_experiment}%
\vspace{-4mm}
\end{figure}

The second experiment is to test the performance of our pose alignment controller. Due to the limitation of point cloud publishing frequency (30~$Hz$), we set the constraints for linear and angular velocities in the Cartesian space as 1.5~$m/s$ and 0.6~$rad/s$ to prevent the substantial deformation of the cable in the interval of the point cloud update. We implement the pose alignment controller with two different cables (10 trials/each). Figures~\ref{fig:pose_alignment_experiment} shows two successful runs during these experiments. Our pose adjustment control loop ends when the pose differences meet the thresholds which is 0.01~$m$ for translational terms and 0.02~$rad$ for rotational terms. In this experiment, we conclude that the HDMI cable (Avg. elapsed time : 12.26~$s$) needs more time than the power cable (Avg. elapsed time : 5.59~$s$) to adjust the pose. A potential reason might be that the HDMI cable has more deformation than the power cable. 

\begin{figure}[ht]
    \centering
    \includegraphics[height=.2\linewidth, width=.3\linewidth]{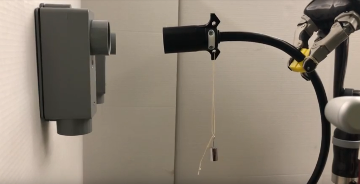}\hfill
    \includegraphics[height=.2\linewidth, width=.3\linewidth]{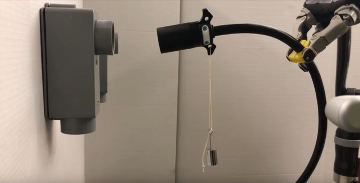}\hfill
    \includegraphics[height=.2\linewidth, width=.3\linewidth]{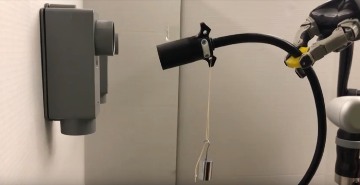}\hfill
    \includegraphics[height=.2\linewidth, width=.3\linewidth]{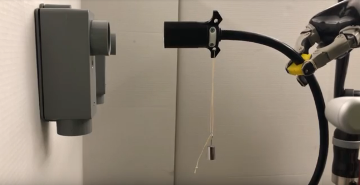}\hfill
    \includegraphics[height=.2\linewidth, width=.3\linewidth]{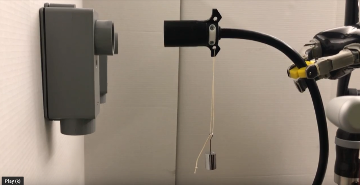}\hfill
    \includegraphics[height=.2\linewidth, width=.3\linewidth]{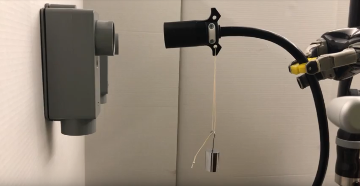}
    \caption{Experiment 3: (top) different weights (20$g$, 50$g$, 100$g$) add to the pose alignment control loop as disturbances, (bottom) final pose of the cable after the pose alignment.}
    \label{fig:disturbance_adding}
\end{figure}

Another experiment is performed for evaluating the robustness of the pose alignment controller by adding disturbances to the pose alignment control loop and testing whether our algorithm can still achieve the target. Figure~\ref{fig:disturbance_adding} shows that different disturbances added to the cable-tip while the system is in the pose alignment loop and the final pose of the cable after the loop, which demonstrates that our pose alignment controller is able to resist such external disturbances.

These results demonstrate that our method for modeling linear flexible objects can provide a 3-D model with adaptive parameters based on the visual feedback. This model is capable of representing various shapes of the cable during the Plug Task. More importantly, our pose alignment algorithm is robust to handle different cable-like objects and is capable of resisting disturbances.

\section{CONCLUSIONS}
In this paper, we have proposed a geometrical modeling method based on the curves on two projection planes for linear flexible objects subject to gravity. The method is capable of tracking the 3-D curvature of the linear flexible object, the pose of the tip and the pose of the selected grasp point on the object. A robust pose alignment controller based on the geometrical model with adaptive parameters can bring the cable-tip to a desired pre-insert position. We have detailed how we formulate and use this method to accomplish the DRC Plug Task autonomously. Moreover, we have performed experiments and demonstrated the versatility, reliability, and robustness of our approach. 

Future work will focus on different model representations of the cable shapes in the case of more complex deformations. Also, we will apply our approach to different robot platforms (e.g., a dual-arm system) or to other flexible object manipulation tasks (e.g., a cable routing assembly task).

\section{Acknowledgements}
This research is supported by the National Aeronautics and Space Administration under Grant No. NNX16AC48A issued through the Science and Technology Mission Directorate, by the National Science Foundation under Award No. 1451427, 1544895, 1928654 and by the Office of the Secretary of Defense under Agreement Number W911NF-17-3-0004.

\bibliographystyle{IEEEtran}
\bibliography{IEEEabrv,IEEEexample}
\end{document}